\documentclass[letterpaper, 10 pt, conference]{ieeeconf}

\IEEEoverridecommandlockouts                              

\usepackage[T1]{fontenc}

\usepackage{amsfonts}	
\usepackage{amsmath}	
\usepackage{amssymb}    
\usepackage{siunitx}
\usepackage{pifont}   
\usepackage{dsfont}   
\usepackage{caption}
\usepackage{booktabs}
\usepackage{makecell}  
\usepackage[flushleft]{threeparttable}  
\usepackage{multirow}
\usepackage{rotating}

\usepackage{xspace}    
\usepackage[dvipsnames]{xcolor}    
\usepackage{colortbl}
\let\labelindent\relax
\usepackage[inline]{enumitem} 
\usepackage{textcomp}  
\usepackage{gensymb}   
\usepackage{graphicx} 
\usepackage[caption=false,font=footnotesize]{subfig}
\usepackage{microtype}
\usepackage{cite}
\usepackage{tikz}
\usepackage{array}

\makeatletter
\let\NAT@parse\undefined
\makeatother

\usepackage{url}

\usepackage[pdfencoding=auto, hidelinks]{hyperref} 
\usepackage[hang,flushmargin]{footmisc}

\usepackage[capitalize]{cleveref}
\crefname{section}{Sec.}{Secs.}
\Crefname{section}{Section}{Sections}
\Crefname{table}{Table}{Tables}
\crefname{table}{Tab.}{Tabs.}

\usepackage{balance}

\newcommand{\net}{\mbox{DualViewMapDet}\xspace}

\definecolor{Gray}{gray}{0.9}

\begin{document}

\title{\LARGE \bf
Leveraging Previous-Traversal Point Cloud Map Priors for Camera-Based 3D Object Detection and Tracking
}

\author{
Markus Käppeler$^{1}$,
Özgün Çiçek$^{2}$,
Yakov Miron$^{2,3}$,
and Abhinav Valada$^{1}$
\thanks{$^{1}$ Department of Computer Science, University of Freiburg, Germany.}%
\thanks{$^{2}$ Bosch Research, Robert Bosch GmbH, Renningen, Germany.}%
\thanks{$^{3}$ University of Haifa, Israel}%
\thanks{This research was funded by Bosch Research as part of a collaboration between Bosch Research and the University of Freiburg on AI-based automated driving.}%
}

\maketitle


\begin{abstract}
    Camera-based 3D object detection and tracking are central to autonomous driving, yet precise 3D object localization remains fundamentally constrained by depth ambiguity when no expensive, depth-rich online LiDAR is available at inference. In many deployments, however, vehicles repeatedly traverse the same environments, making static point cloud maps from prior traversals a practical source of geometric priors. We propose \net, a camera-only inference framework that retrieves such map priors online and leverages them to mitigate the absence of a LiDAR sensor during deployment. The key idea is a dual-space camera–map fusion strategy that avoids one-sided view conversion. Specifically, we (i) project the map into perspective view (PV) and encode multi-channel geometric cues to enrich image features and support BEV lifting, and (ii) encode the map directly in bird’s-eye view (BEV) with a sparse voxel backbone and fuse it with lifted camera features in a shared metric space. Extensive evaluations on nuScenes and Argoverse~2 demonstrate consistent improvements over strong camera-only baselines, with particularly strong gains in object localization. Ablations further validate the contributions of PV/BEV fusion and prior-map coverage.  We make the code and pre-trained models available at \mbox{\url{https://dualviewmapdet.cs.uni-freiburg.de}}. 
\end{abstract}


\section{Introduction}

\label{sec:introduction}

Camera-based 3D object detection and tracking is a core perception capability for autonomous driving, enabling the localization and tracking of dynamic agents for downstream prediction and planning. Many recent driving systems rely on object-centric scene representations, often combined with sparse maps, as an interface for interaction modeling and decision making~\cite{sun2024sparsedrive}. While LiDAR-only~\cite{yin2021center, lang2024point} and camera-LiDAR fusion methods~\cite{liu2022bevfusion, cai2023bevfusion4d, yin2024fusion, mohan2024progressive} achieve impressive performance leveraging precise metric depth and 3D geometry, they require expensive LiDARs at inference that limit scalability. This motivates camera-only 3D detection~\cite{lin2023sparse4dv3, kappeler2025bridging}, which uses inexpensive multi-view cameras that provide rich semantic cues, but remains fundamentally challenged by depth ambiguity. Errors in depth estimation directly degrade 3D box localization~\cite{li2023bevdepth}. This limitation affects both methods operating in the \emph{perspective view} (PV)~\cite{liu2022petr,wang2023exploring,lin2023sparse4dv3,jiang2024far3d,liu2024ray} and approaches that lift image features into a \emph{bird's-eye view} (BEV)~\cite{yang2023bevformer,philion2020lift,li2023bevdepth,li2024bevnext}, since reliable BEV lifting requires accurate depth estimates.

\begin{figure}[t]
    \centering
    \includegraphics[width=\linewidth]{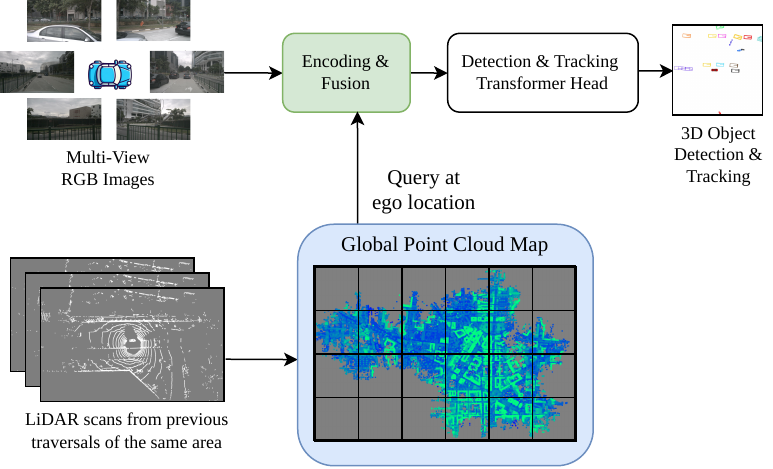}
    \caption{Overview of our camera-map fusion framework. Given multi-view camera images and the current ego location, we retrieve a local static LiDAR point cloud map constructed from previous mapping traversals of the same area. The map is encoded in both perspective view (PV) and bird's-eye view (BEV) and fused with the corresponding image features in each representation. A transformer-based head then predicts 3D bounding boxes and object tracks from the fused PV/BEV features.}
    \label{fig:cover}
    \vspace{-0.4cm}
\end{figure}

\begin{figure}[t]
    \centering
    \includegraphics[width=0.85\linewidth]{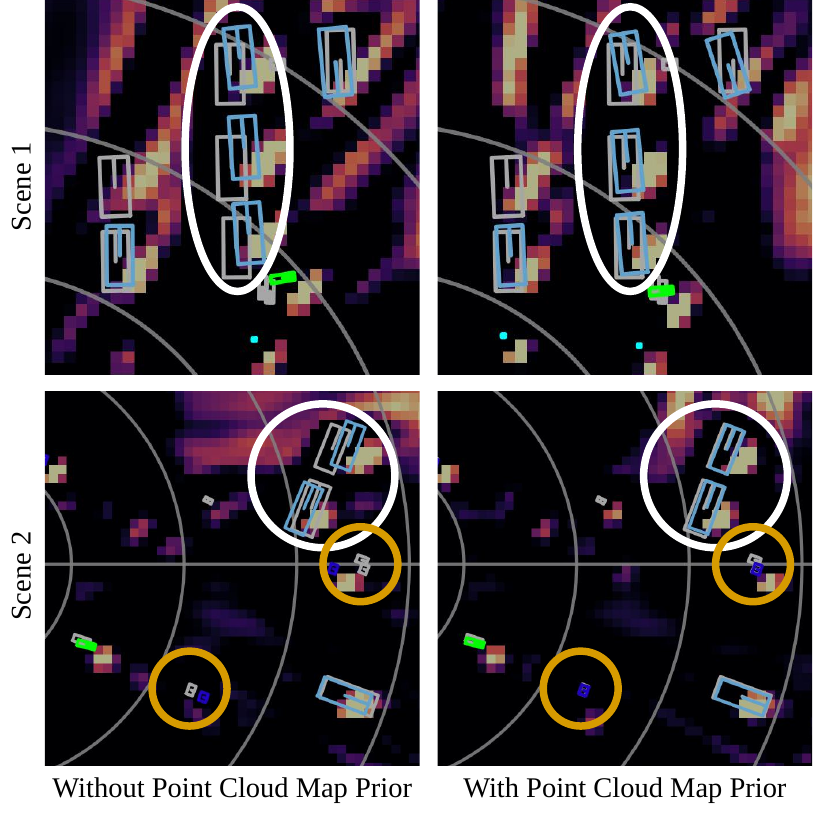}
    \caption{Map priors reduce depth ambiguity. Compared to purely image-based 3D object detection (left), incorporating a static point cloud map provides accurate geometric anchors in the scene, improving 3D box localization for dynamic objects (right). Top-down BEV detections of cars are shown in light blue, detections of pedestrians in dark blue, and ground truth in gray. The ego car is at the center of the circles. In the background, we visualize BEV activations from a single channel that responds strongly to cars. Activations with a map prior are less distorted along the depth.}
    \label{fig:motivation}
    \vspace{-0.3cm}
\end{figure}

A promising direction to reduce depth ambiguity is to incorporate geometric priors beyond the current observation. In many real-world deployments, vehicles operate in environments that have been traversed before by a mapping vehicle, making prior maps available. In particular, 3D LiDAR point cloud maps constructed from earlier traversals provide accurate geometry of the static surroundings at the same metric scale as the detection targets. Such priors can anchor object depth relative to nearby static structure (e.g., road surface, curbs, signs) and provide spatial context that helps disambiguate object localization even when objects are far away or partially occluded (Fig.~\ref{fig:motivation}). Importantly, unlike camera-LiDAR fusion, these maps encode only the \emph{static} environment and thus must be exploited as a prior rather than as a source of current dynamic measurements.

In this paper, we show that a prior point cloud map of the static environment, which is retrieved at the ego location, can substantially improve camera-based 3D object detection and tracking while preserving camera-only inference. Fig.~\ref{fig:cover} shows an overview of our camera–map fusion framework. We address the representation gap between sparse 3D geometry and dense multi-view image features with a dual-space camera-map fusion architecture. First, we project the retrieved map onto each camera and construct multi-channel PV map encodings that include sparse depth, depth positional embeddings, and ego-frame point coordinates. This makes image features explicitly depth-aware for the head and eases depth estimation for BEV lifting. Second, we encode the map directly in BEV using a sparse voxel backbone and fuse it with lifted camera features in a shared metric BEV space, which we find crucial for fully exploiting map priors to achieve accurate object localization. Our approach, \net, consistently improves strong camera-only methods on large-scale benchmarks such as nuScenes~\cite{caesar2020nuscenes} and Argoverse~2~\cite{wilson2023argoverse}, with particularly strong improvements in object translation accuracy, highlighting the value of map priors for precise 3D perception without requiring LiDAR at test time.

To summarize, our main contributions are:
\begin{enumerate}[topsep=0pt]
    \item A camera-only inference framework that leverages prior point cloud maps from previous traversals to reduce depth ambiguity and improve 3D object localization, without requiring LiDAR at test time.
    \item We propose multi-channel \emph{perspective-view} map encodings of the projected point cloud prior (e.g., sparse depth with validity masks, depth positional embeddings, and ego-frame geometry cues) to inject geometric priors into image features, supporting depth estimation and detection.
    \item We design, for the first time, a dual-space fusion mechanism that combines camera and map information in PV, and additionally fuses lifted camera BEV features with a sparse voxel-encoded BEV map prior in a shared metric BEV space. This enables effective exploitation of map geometry for accurate 3D object detection and resolving design decisions in autonomous driving architectures, i.e., how to use map priors and how the data flow should be from PV and BEV views.
    \item A image–map grid masking strategy, which further improves robustness to missing or imperfect map priors.
\end{enumerate}

\section{Related Work}
\label{sec:related-work}


We review prior work on 3D object detection and the use of geometric priors from maps of previous traversals, focusing on camera-based detection, multi-modal fusion, and map priors to improve 3D perception tasks.
\subsection{3D Object Detection}
3D object detection predicts 3D bounding boxes from sensor data such as multi-view cameras, LiDAR, or radar. Key challenges include multi-view fusion, temporal modeling, and multi-modal fusion across heterogeneous data representations.

{\parskip=2pt
\noindent\textit{LiDAR-Based Detection:}
LiDAR provides accurate geometry and depth, but is sparse at long range. Voxel-based detectors encode point clouds using sparse 3D convolutions, as popularized by SECOND~\cite{yan2018second}, and many approaches detect objects from BEV features such as CenterPoint~\cite{yin2021center}. These designs highlight the effectiveness of BEV as a metric representation for detection and motivate our use of a sparse voxel-based BEV encoder to encode the point cloud map prior.

{\parskip=2pt
\noindent\textit{Camera-Based Detection:}
Camera-only methods provide rich semantic and long-range cues, but suffer from depth ambiguity~\cite{nallapareddy2023evcenternet}. Most approaches fall into dense BEV-based methods that lift image features into a unified BEV grid~\cite{yang2023bevformer,philion2020lift,li2023bevdepth,li2024bevnext} and sparse query-based methods that refine a set of 3D object queries by cross-attending to multi-view image features~\cite{liu2022petr,wang2023exploring,lin2023sparse4dv3,jiang2024far3d,liu2024ray}. BEV-based methods use attention-based view transformation (e.g., BEVFormer~\cite{yang2023bevformer}) or LSS-style lifting~\cite{philion2020lift,li2024bevnext}, where depth estimation is the bottleneck. Thus, supervising depth with LiDAR improves lifting and accuracy~\cite{li2023bevdepth}. Query-based approaches avoid dense BEV features by directly aggregating image features at projected 3D reference points~\cite{liu2022petr,jiang2024far3d, liu2024ray} and often perform temporal query propagation~\cite{wang2023exploring,lin2023sparse4dv3}. We build on the strong query-based head design of Sparse4Dv3~\cite{lin2023sparse4dv3}, but explicitly inject geometric priors from a point cloud map to address depth ambiguity, and fuse these map priors in both PV and BEV to leverage complementary representations.

{\parskip=2pt
\noindent\textit{Camera--LiDAR Fusion:}
Camera-LiDAR fusion combines complementary strengths~\cite{buchner20223d} where cameras provide semantics and high-resolution cues, while LiDAR provides metric geometry. Many methods fuse in BEV as a shared metric representation. TransFusion~\cite{bai2022transfusion} injects image cues into LiDAR query-proposals via cross-attention, while BEVFusion~\cite{liu2022bevfusion} lifts images to BEV and fuses them with LiDAR BEV features before box decoding. Subsequent work improves view transformation and fusion robustness, often using LiDAR to guide lifting and alignment~\cite{cai2023bevfusion4d,hu2023ea,yin2024fusion,song2024graphbev}. While highly effective, these methods require LiDAR sensor data at inference. In contrast, we target camera-only inference while leveraging a \emph{prior} point cloud map from previous traversals. We enhance PV image features with multiple map-derived encodings to support both the detection head and the LSS depth network, and additionally fuse map features with camera features in BEV to better exploit metric structure.

\noindent\textit{3D Multi-Object Tracking:} Most modern 3D multi-object tracking approaches are built on top of strong 3D detectors, often by reusing their box predictions and adding temporal association or propagation modules~\cite{gosala2026sparse3dtrack, lin2023sparse4dv3, wang2023exploring}.

\subsection{Geometric Prior Maps for 3D Perception}

{\parskip=0pt
\noindent\textit{Prior Maps for 3D Occupancy:}
Several works use priors from previous traversals for occupancy prediction, e.g., NeRF-based maps~\cite{yuan2024presight} and global occupancy maps~\cite{yuan2025lmpocc}. However, occupancy is voxelized and coarse, whereas 3D object detection requires precise continuous object localization.

{\parskip=2pt
\noindent\textit{Multi-Task Geometry for Detection:}
Other approaches couple detection and occupancy estimation to enrich geometric supervision and fine-grained spatial structure~\cite{yuan2025collaborative,yu2024ultimatedo}. Although beneficial, these methods are not primarily optimized for state-of-the-art 3D detection accuracy.

{\parskip=2pt
\noindent\textit{Map Priors for Detection:}
Beyond occupancy, prior traversals have been exploited to build maps that support detection, including multi-traversal consensus to detect dynamic content in reconstructed scene representations~\cite{li2024memorize} and reusing LiDAR maps for LiDAR-based online detection~\cite{you2022hindsight}. Most closely related, AsyncDepth~\cite{you2024better} improves camera-based 3D detection by projecting an aggregated historical point cloud into per-camera depth maps and concatenating them with image features in PV. However, PV-only fusion is limited by the representational gap between sparse 3D geometry and image features (see \cref{tab:ablation-bev-map}), and does not explicitly leverage BEV as a metric fusion space.

\section{Technical Approach}
\label{sec:method}

\begin{figure*}[t]
    \centering
    \includegraphics[width=\linewidth]{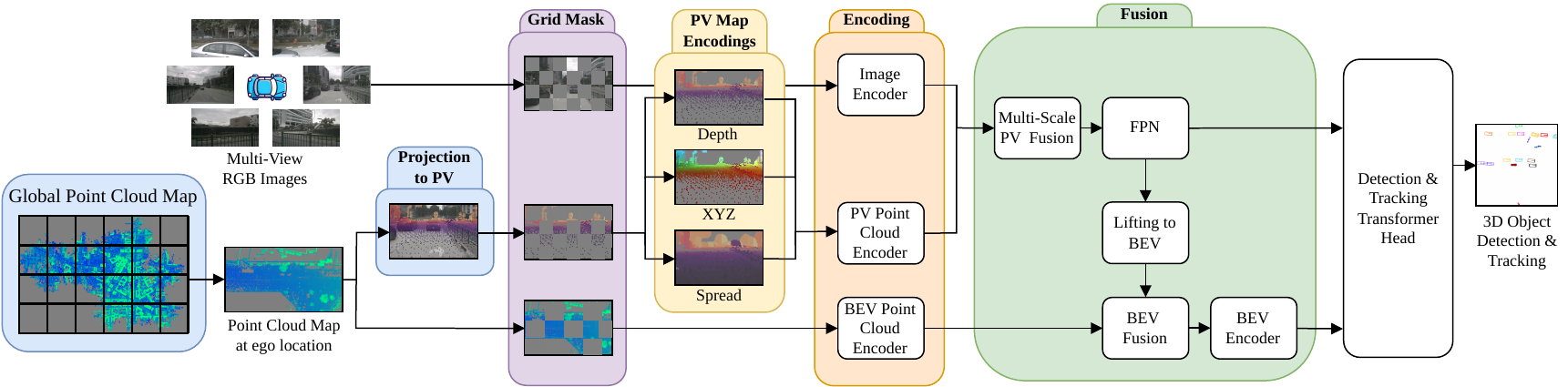}
    \caption{
Overview of \net. Given multi-view RGB images and a local static point cloud map patch retrieved at the current ego location, we employ image-map grid masking during training to improve robustness (omitted in PV map encodings). The images are encoded into multi-scale PV features, while the map is encoded in \emph{perspective view} by projecting it into each camera to form multi-channel PV map encodings and in \emph{bird's-eye view} into a BEV feature grid. We then fuse camera and map features in PV, lift the fused PV features to BEV, and fuse them with the BEV map features in a shared metric space. Finally, a transformer-based head with BEV-PV deformable aggregation decodes 3D bounding boxes from the fused PV and BEV representations.
}
    \label{fig:overview}
    \vspace{-0.3cm}
\end{figure*}

We propose \net, a camera-only framework for 3D object detection and tracking augmented with an offline static point cloud \emph{prior map} from previous traversals (Fig.~\ref{fig:overview}). At inference, \net requires only multi-view RGB images, calibrated cameras, a globally localized ego pose, and an online retrieval of a local static map patch. \emph{No live LiDAR sensor} is used. The central challenge is to exploit a 3D geometric map prior for precise 3D box localization while preserving the semantic richness of image features.\looseness=-1

A static point cloud map can anchor depth and provide metric context, but using it effectively is non-trivial due to the mismatch between sparse 3D geometry and dense image features. \emph{PV-only} fusion lacks an explicit metric space where nearby pixels may correspond to distant 3D locations, limiting geometric context aggregation, particularly under occlusions. \emph{BEV-only} fusion enables metric reasoning but can lose fine PV semantics during view transformation and does not directly assist the depth estimation required by LSS-based lifting~\cite{philion2020lift}. We therefore adopt a \emph{dual-space camera-map fusion} architecture:
(i) we project the map into PV to build multi-channel geometric encodings aligned with each image, making PV features explicitly depth-aware and easing depth estimation for lifting; (ii) we also encode the map in BEV using a sparse voxel backbone to exploit metric structure for localization. Then, we fuse the map features projected to PV with camera features in PV, and lift it to BEV. (iii) Finally, we fuse the lifted fused features in BEV with the map encoded in BEV, and decode boxes with a transformer head whose queries sequentially attend to fused BEV and fused PV features. 


\subsection{Problem Setup}
At time $t$, we observe synchronized surround-view images $\{I_t^c\}_{c=1}^{N}$ with known intrinsics $K_c$ and extrinsics $T_{E\rightarrow C_c}$ (ego-to-camera). We assume global ego localization $T^{G}_{E_t}\in SE(3)$. We are given an offline-built global \emph{static} point cloud map $\mathcal{M}^{G}=\{\mathbf{p}^{G}\}$ from previous traversals that contains only static structure. The goal is to predict a set of oriented 3D boxes in the ego frame:
\begin{equation}
\mathcal{B}_t = \{ b_i^t \}_{i=1}^{M},\qquad
b_i^t = (x,y,z,w,h,l,\mathrm{yaw},\mathbf{v}),
\end{equation}
where $(x,y,z)$ is the center, $(w,h,l)$ are dimensions, and $\mathbf{v}$ denotes velocity.


\subsection{Prior Map Construction and Online Retrieval}
\label{ssec:method-map}
We build $\mathcal{M}^{G}$ offline by accumulating LiDAR sweeps from \emph{other} traversals and removing dynamic points using 3D boxes. To enable efficient retrieval, we store the map in spatial tiles of size $L\times L$ meters. At inference, given ego translation $\mathbf{t}^{G}_{E_t}$ and perception range $R$, we retrieve a local patch
\begin{equation}
\mathcal{P}_t^{G} = \left\{\mathbf{p}^{G}\in \mathcal{M}^{G}\ \middle|\ \lVert (p_x^{G},p_y^{G}) - (t_{x}^{G},t_{y}^{G})\rVert_{\infty}\le R \right\}.
\end{equation}
To avoid leakage, we exclude points originating from the current traversal and traversals too close in time. We merge retrieved points, remove noise via statistical outlier removal~\cite{zhou2018open3d}, and downsample by average voxel pooling with voxel size $\SI{0.4}{\meter}$ for efficiency. Finally, we transform the patch into the current ego frame:
\begin{equation}
\mathcal{P}_t^{E}=\left\{\mathbf{p}^{E}=(T^{G}_{E_t})^{-1}\mathbf{p}^{G}\ \middle|\ \mathbf{p}^{G}\in \mathcal{P}_t^{G}\right\}.
\end{equation}
This ego-frame static map prior $\mathcal{P}_t^{E}$ is the only map input to \net at test time.


\subsection{Dual-Space Map Encoding}
\label{ssec:method-encoding}
We bridge sparse 3D map geometry and dense image features by encoding the retrieved map in \emph{both} PV and BEV.

\subsubsection{Multi-view Image Features}
We extract multi-scale PV features from each camera image using an image backbone:
\begin{equation}
\mathbf{F}^{c,s}_{\text{img}} = \phi_{\text{img}}^{s}(I_t^c),\qquad s\in\mathcal{S},
\end{equation}
where $\mathcal{S}$ denotes backbone feature scales.

\subsubsection{PV Map Rasterization and Multi-Channel Encodings}
\label{ssec:method-pv-encoding}

We project the retrieved ego-frame map patch $\mathcal{P}_t^{E}$ into each camera using the known intrinsics and extrinsics to obtain a sparse depth map and additional geometric cues aligned with the image plane. We rasterize a sparse depth image $D_t^c\in\mathbb{R}^{H\times W}$ using a z-buffer rule: if multiple projected points fall into the same pixel, we keep the closest depth. We also build a validity mask $M_t^c\in\{0,1\}^{H\times W}$, where $M_t^c(u,v)=1$ iff at least one map point projects to pixel $(u,v)$. Since the prior map is sparse, this mask enables the network to distinguish reliable geometric anchors from uncovered pixels.

\paragraph*{Ego-frame coordinate channels}
For pixels with a valid point, we store its ego-frame coordinates $\mathbf{p}^{E}=(x,y,z)$ as three additional channels normalized by the perception range. This makes PV features explicitly aware of metric 3D structure and the ego-centric coordinate system in which boxes are regressed, providing a direct cue for localization beyond depth alone (e.g., height via $z$).

\paragraph*{Depth positional embedding}
We embed normalized depth $d_{\text{norm}}=d/d_{\max}$, where $d=D_t^c(u,v)$, via a sinusoidal mapping (positional encoding)~\cite{vaswani2017attention,cattaneo2025cmrnext} into $\Phi(d_{\text{norm}})$.

\paragraph*{Nearest-valid depth spread and distance-to-valid}
Sparse depth leaves large uncovered regions. We therefore add two channels that (i) spread the nearest valid depth within a bounded pixel radius $r$ and (ii) encode uncertainty via the (clipped and normalized) distance to the nearest valid pixel. Concretely, for each pixel we copy the depth value $d_{\text{norm}}$ of the nearest valid pixel to $d_{\text{near}}$ if it lies within radius $r$ (otherwise we set the spread depth to zero), and we additionally store the normalized distance-to-valid $\delta_{\text{norm}} \in [0,1]$.

\paragraph*{Final PV map tensor and encoder}
We concatenate all cues to form a multi-channel PV map tensor:
\begin{equation}
\mathbf{X}_t^c(u,v)=\Big[d_{\text{norm}},\ M_t^c,\ \Phi(d_{\text{norm}}),\ \mathbf{p}^{E}_{\text{norm}},\ d_{\text{near}},\ \delta_{\text{norm}}\Big] \in \mathbb{R}^{C_m},
\end{equation}
and encode it with a lightweight CNN (ResNet-18~\cite{he2016deep}) into multi-scale PV map features:
\begin{equation}
\mathbf{F}^{c,s}_{\text{map}}=\phi^{s}_{\text{pvmap}}(\mathbf{X}_t^c).
\end{equation}

\subsubsection{BEV Map Encoding via Sparse Voxels}
We additionally encode the retrieved map $\mathcal{P}_t^{E}$ directly in BEV using a sparse voxel backbone~\cite{yan2018second}. We use a voxel size of $0.2\,$m and compute per-voxel features as the mean 3D coordinate (mean $x,y,z$) of points within the voxel. A sparse 3D convolutional encoder then produces a dense BEV feature grid:
\begin{equation}
\mathbf{F}_{\text{BEV}}^{\text{map}}
=
\phi_{\text{bevmap}}\!\left(\mathrm{Voxelize}(\mathcal{P}_t^{E})\right)
\in \mathbb{R}^{Y\times X\times C_b}.
\end{equation}


\subsection{Dual-Space Camera-Map Fusion}
\label{ssec:method-fusion}
We fuse camera and map information in PV to support depth reasoning and PV decoding, then fuse again in BEV to exploit metric structure.

\subsubsection{PV Fusion and Post-Fusion FPN}
At each backbone scale $s$, we fuse image and PV map features by concatenating along the channel dimension and applying a $3\times3$ Conv (with BN and ReLU) to reduce channels:
\begin{equation}
\mathbf{F}^{c,s}_{\text{PV}}
=
\psi_{\text{PV}}^{s}\!\left(\left[\mathbf{F}^{c,s}_{\text{img}} \ \Vert\ \mathbf{F}^{c,s}_{\text{map}}\right]\right),
\end{equation}
where $[\cdot\Vert\cdot]$ denotes channel-wise concatenation and $\psi_{\text{PV}}^{s}$ denotes the $3\times3$ Conv block. We then employ an FPN on top of $\{\mathbf{F}^{c,s}_{\text{PV}}\}_{s\in\mathcal{S}}$ to fuse multi-scale information and obtain the final fused PV pyramid used by the head and the BEV lifting module.

\subsubsection{Depth-Aware BEV Lifting}
We lift fused PV features into BEV using an LSS-style module~\cite{li2024bevnext, philion2020lift}. The module predicts a per-pixel discrete depth distribution $\mathbf{P}_{\text{depth}}^c(u,v)\in[0,1]^D$ and forms frustum features via an outer product:
\begin{equation}
\mathbf{F}^c_{\text{frustum}}(u,v)
=
\mathbf{P}_{\text{depth}}^c(u,v)\otimes \mathbf{F}^{c}_{\text{PV}}(u,v)
\in \mathbb{R}^{D\times C}.
\end{equation}
Unprojected frustum features from all cameras are pooled into a BEV grid, yielding $\mathbf{F}_{\text{BEV}}^{\text{cam}}$. Using fused PV features makes depth estimation easier as the map provides accurate geometric anchors in static regions, thereby reducing lifting ambiguity for nearby dynamic objects (via surrounding context) and improving BEV alignment.

\subsubsection{BEV Fusion}
We fuse lifted camera BEV features with BEV map prior features in the shared metric space by concatenating along the channel dimension and applying a $3\times3$ Conv (with BN and ReLU):
\begin{equation}
\mathbf{F}_{\text{BEV}}
=
\psi_{\text{BEV}}\!\left(\left[\mathbf{F}_{\text{BEV}}^{\text{cam}} \ \Vert\ \mathbf{F}_{\text{BEV}}^{\text{map}}\right]\right),
\end{equation}
followed by a lightweight BEV encoder (ResNet blocks + small FPN) to refine and distribute fused context. This BEV fusion is critical for precise 3D object localization because it enables reasoning in a metric space where distances and neighborhood relations correspond to 3D geometry—something PV-only fusion cannot provide.


\subsection{Detection Head with BEV-PV Deformable Aggregation}
\label{ssec:method-head}
We adopt the detection head from DualViewDistill~\cite{kappeler2025bridging}, which builds upon Sparse4Dv3~\cite{lin2023sparse4dv3}. The head maintains $M$ object queries consisting of (i) 3D anchors and (ii) instance features:
\begin{equation}
A\in\mathbb{R}^{M\times 11},\qquad F\in\mathbb{R}^{M\times C},
\end{equation}
where each anchor encodes center, dimensions, yaw (as $\sin/\cos$), and velocity:
\begin{equation}\small
\textbf{a}_i=(x,y,z,\ln w,\ln h,\ln l,\sin\mathrm{yaw},\cos\mathrm{yaw},v_x,v_y,v_z).
\end{equation}

\paragraph*{Sequential BEV then PV aggregation.}
In each decoder layer, queries $\textbf{f}_i$ interact with \emph{both} BEV and PV features via deformable aggregation over a set of 3D keypoints sampled on the anchor box.
Each keypoint is (i) projected into the BEV grid to bilinearly sample $\mathbf{F}_{\text{BEV}}$, and (ii) projected into each camera to bilinearly sample multi-scale fused PV features $\{\mathbf{F}^{c,s}_{\text{PV}}\}$. We fuse modalities sequentially:
\begin{equation}
\begin{aligned}
\mathbf{f}_i'  &= \mathbf{f}_i + \text{BEV-DeformAgg}\!\left(\mathbf{f}_i, \mathbf{a}_i, \mathbf{F}_{\text{BEV}}\right), \\
\mathbf{f}_i'' &= \mathbf{f}_i' + \text{PV-DeformAgg}\!\left(\mathbf{f}_i', \mathbf{a}_i, \{\mathbf{F}^{c,s}_{\text{PV}}\}\right),
\end{aligned}
\end{equation}
where BEV aggregation injects metric context (including the BEV map prior) and PV aggregation injects fine-grained semantic details from fused PV features. This ordering aligns with our motivation: BEV provides spatially grounded cues for localization, while PV refines object-centric semantics and details that may be attenuated in BEV.


\subsection{Image--Map Grid Masking}
\label{ssec:method-mask-modeling}
To improve robustness to missing or imperfect priors and avoid overfitting to map cues, we employ grid masking augmentation~\cite{chen2020gridmask} to (i) RGB images, (ii) PV map tensors $\mathbf{X}_t^c$, and (iii) the BEV map prior by removing all map points whose BEV coordinates fall into masked ground regions. This improves robustness to missing or imperfect map priors and avoids overfitting to the map.


\subsection{Training Objective}
\label{ssec:method-training}
We train \net end-to-end. The map prior is learned to be exploited \emph{only through detection gradients}. The total loss is
\begin{equation}
\mathcal{L}_{\text{total}} = \lambda_{\text{det}}\mathcal{L}_{\text{det}} + \lambda_{\text{depth}}\mathcal{L}_{\text{depth}},
\end{equation}
where $\mathcal{L}_{\text{det}}$ is the standard detection loss of the Sparse4D head (classification + box regression + orientation/auxiliary terms)~\cite{lin2023sparse4dv3}. $\mathcal{L}_{\text{depth}}$ supervises the LSS depth estimator with LiDAR depth labels~\cite{li2023bevdepth, li2024bevnext} during training, while inference remains camera-only sensing augmented with the retrieved static prior map.

\section{Experimental Evaluation}
\label{sec:result}

We evaluate \net on large-scale autonomous driving datasets to quantify the benefit of prior point cloud maps for camera-based 3D object detection and tracking. Our experiments are designed to (i) compare against strong camera-only baselines under standard dataset evaluation protocols, and (ii) isolate key design choices in map encoding, fusion location (PV vs.\ BEV), and robustness to incomplete or missing map priors. Unless noted otherwise, we report the official dataset metrics and follow common training and evaluation conventions to ensure comparability.


\subsection{Implementation and Training Details}

Both the image backbone and the PV map encoder produce multi-scale features at scales $\{4,8,16,32\}$. For the BEV branch, we use a BEV grid of $128\times128$. BEV features from the map encoder and the lifted PV features share the same spatial resolution. For BEV lifting, we lift stride-$8$ PV features for smaller input resolutions (nuScenes and Argoverse~2 ablations) and stride-$16$ PV features for the higher-resolution Argoverse~2 model. On Argoverse~2, we use VoVNet-99~\cite{lee2019energy} initialized from FCOS3D pretraining on nuScenes, and ResNet-50~\cite{he2016deep} initialized from ImageNet, following our baselines. The PV map encoder is a lightweight ResNet-18 initialized from ImageNet. For optimization, we train all models for 80 epochs with AdamW, using cosine learning-rate decay with a 500-iteration warm-up and a base learning rate of $2\times10^{-4}$. We use a batch size of 24 and train with a sequential iteration strategy. We apply both image-level and BEV/3D augmentations as in~\cite{kappeler2025bridging}, and supervise the LSS depth estimator with LiDAR depth during training while keeping inference camera-only. For Argoverse~2 long-range evaluation at $\SI{150}{\meter}$, we increase the number of object queries from 900 to 1800 to cover the larger range. To reduce memory usage, we keep the BEV grid range at $\SI{50}{\meter}$ and rely on PV features beyond the BEV extent.


\subsection{Datasets}

We evaluate on nuScenes~\cite{caesar2020nuscenes} and Argoverse~2~\cite{wilson2023argoverse}, which provide multi-view surround camera rigs and LiDAR sensor data. Both datasets contain repeated coverage of urban areas across many sequences, enabling the construction of prior point cloud maps from other sequences. To avoid any leakage from evaluation data, we construct training maps exclusively from the training split. For evaluation, we use maps from the training and validation splits, but exclude the map of the current traversal. For \textit{nuScenes}, we follow the official protocol and report NDS and mAP as the primary detection metrics, together with the true-positive errors (mATE, mASE, mAOE, mAVE, mAAE). For multi-object tracking, we report AMOTA as the primary metric and the standard tracking metrics. For \textit{Argoverse~2}, we report the official Composite Detection Score (CDS) and mAP, together with the standard true-positive errors (mATE, mASE, mAOE) across the 26 object categories.


\subsection{Global Ego Pose Correction}

Accurate global alignment across traversals is crucial for map retrieval. Even small cross-sequence offsets can corrupt the projected PV depth cues and degrade BEV fusion. Argoverse~2 provides 6-DoF ego poses within each sequence, but we observe cross-sequence global misalignments of up to a few meters. We therefore refine global poses by performing point cloud matching between overlapping sequences using KISS-Matcher~\cite{lim2025kiss} and solving a pose graph optimization to obtain a globally geometrically consistent map prior. We will provide the corrected poses for future research. In contrast, nuScenes provides global poses with a fixed height component (effectively 5-DoF), which introduces errors when projecting map points into PV on non-flat roads. Since this limitation is difficult to fix, we use the provided poses as-is for nuScenes and emphasize Argoverse~2 as our primary testbed. 


\subsection{3D Object Detection and Tracking}

\begin{table}

\begin{minipage}{\linewidth}
\footnotesize
\centering
\caption{Long-range 3D object detection results on Argoverse 2 val set.}
\label{tab:results-argoverse2-150m}
\setlength\tabcolsep{2.5pt}
\begin{threeparttable}
    \begin{tabular}{l | c c | c c c c }
        \toprule
        \textbf{Method}  & \textbf{CDS$\uparrow$} & \textbf{mAP$\uparrow$} & \textbf{mATE$\downarrow$} & \textbf{mASE$\downarrow$} & \textbf{mAOE$\downarrow$} \\
        \midrule
        PETR\textsuperscript{*}~\cite{liu2022petr} & 0.122 & 0.176 & 0.911 & 0.339 & 0.819 \\
        StreamPETR\textsuperscript{*}~\cite{wang2023exploring} & 0.146 & 0.203 & 0.843 & 0.321 & 0.650 \\
        RayDN\textsuperscript{*}~\cite{liu2024ray} & 0.161 & 0.223 & 0.825 & 0.325 & 0.629 \\ 
        Far3D\textsuperscript{*}~\cite{jiang2024far3d} & 0.181 & 0.244 & 0.796 & 0.304 & 0.538 \\
        Sparse4Dv3\textsuperscript{*}\textsuperscript{\textdaggerdbl}~\cite{lin2023sparse4dv3} & 0.186 & 0.247 & 0.692 & 0.302 & 0.558 \\
        Sparse4Dv3\textsuperscript{\S}\textsuperscript{\textdaggerdbl}~\cite{lin2023sparse4dv3} & 0.182 & 0.244 & 0.726 & 0.307 & 0.548 \\
        \rowcolor{Gray} \net\textsuperscript{\S} & \textbf{0.202} & \textbf{0.263} & \textbf{0.667} & \textbf{0.291} & \textbf{0.515} \\
        \bottomrule
    \end{tabular}
    \footnotesize
    We evaluate within a range of 150 meters.
\end{threeparttable}
\vspace{+0.3cm}
\end{minipage}

\begin{minipage}{\linewidth}
\footnotesize
\centering
\caption{Near-range 3D object detection results on Argoverse 2 val set.}
\label{tab:results-argoverse2-50m}
\setlength\tabcolsep{2.5pt}
\begin{threeparttable}
    \begin{tabular}{l | c c | c c c c }
        \toprule
        \textbf{Method} & \textbf{CDS$\uparrow$} & \textbf{mAP$\uparrow$} & \textbf{mATE$\downarrow$} & \textbf{mASE$\downarrow$} & \textbf{mAOE$\downarrow$} \\
        \midrule
        Far3D\textsuperscript{*}~\cite{jiang2024far3d} & 0.281 & 0.367 & 0.730 & 0.300 & 0.531 \\
        Sparse4Dv3\textsuperscript{*}\textsuperscript{\textdaggerdbl}~\cite{lin2023sparse4dv3} & 0.294 & 0.381 & 0.706 & 0.314 & 0.666 \\
        Sparse4Dv3\textsuperscript{\S}\textsuperscript{\textdaggerdbl}~\cite{lin2023sparse4dv3} & 0.288 & 0.376 & 0.694 & 0.298 & 0.547 \\
        \rowcolor{Gray} \net\textsuperscript{\S} & \textbf{0.311} & \textbf{0.397} & \textbf{0.622} & \textbf{0.279} & \textbf{0.527} \\
        \bottomrule
    \end{tabular}
    \footnotesize
    We evaluate within a range of 50 meters.
\end{threeparttable}
\vspace{+0.3cm}
\end{minipage}

\centering
\begin{minipage}{\linewidth}
\footnotesize
\cref{tab:results-argoverse2-150m} and \cref{tab:results-argoverse2-50m}: All methods use VoVNet-99 as the image encoder with an input image resolution of $640 \times 960$.
\textsuperscript{*}:~Training uses the $\SI{10}{\hertz}$ Argoverse~2 sequences by splitting each sequence into five offset subsequences, yielding \(\approx 5\times\) more (but redundant) training samples than strict $\SI{2}{\hertz}$ subsampling.
\textsuperscript{\S}:~Training with strict $\SI{2}{\hertz}$ subsampling.
\textsuperscript{\textdaggerdbl}:~Baselines trained by us with the code provided by the authors.
\end{minipage}
\vspace{-0.4cm}

\end{table}

\begin{table}
\footnotesize
\centering
\caption{3D object detection results on the nuScenes val set.}
\label{tab:results-3d-val}
\setlength\tabcolsep{0.2pt}
\begin{threeparttable}
    \begin{tabular}{l | c c | c c c c c}
        \toprule
        \textbf{Method} & \textbf{NDS$\uparrow$} & \textbf{mAP$\uparrow$} & \textbf{mATE$\downarrow$} & \textbf{mASE$\downarrow$} & \textbf{mAOE$\downarrow$} & \textbf{mAVE$\downarrow$} & \textbf{mAAE$\downarrow$} \\
        \midrule
        BEVFormerv2~\cite{yang2023bevformer}\textsuperscript{\textdagger} & 0.529 & 0.423 & 0.618 & 0.273 & 0.413 & 0.333 & 0.188 \\
        VideoBEV~\cite{han2024exploring} & 0.535 & 0.422 & 0.564 & 0.276 & 0.440 & 0.286 & 0.198 \\
        StreamPETR~\cite{wang2023exploring} & 0.540 & 0.432 & 0.581 & 0.272 & 0.413 & 0.295 & 0.196 \\
        Sparse4Dv2 & 0.539 & 0.439 & 0.598 & 0.270 & 0.475 & 0.282 & 0.179 \\
        BEVNeXt~\cite{li2024bevnext} & 0.548 & 0.437 & 0.550 & 0.265 & 0.427 & 0.260 & 0.208 \\
        Sparse4Dv3\textsuperscript{*}~\cite{lin2023sparse4dv3} & 0.561 & 0.469 & 0.553 & 0.274 & 0.476 & 0.227 & 0.200 \\
        Sparse4Dv3\textsuperscript{\textdaggerdbl}~\cite{lin2023sparse4dv3} & 0.553 & 0.447 & 0.541 & \textbf{0.262} & 0.500 & 0.225 & 0.179 \\
        \rowcolor{Gray} \net & \textbf{0.588} & \textbf{0.481} & \textbf{0.484} & \textbf{0.262} & \textbf{0.393} & \textbf{0.212} & \textbf{0.176} \\
        \bottomrule
    \end{tabular}
    \footnotesize
    All methods use ResNet-50 as the image encoder with an input image resolution of  $256 \times 704$.
    \textdagger:~Uses pre-trained weights from the nuImage dataset.
    \textdaggerdbl:~Baselines trained with the code provided by the authors.
    *:~Results reported in the original paper but not reproducible.
\end{threeparttable}
\vspace{-0.3cm}
\end{table}

\begin{table}
\footnotesize
\centering
\caption{3D multi-object tracking results on the nuScenes val set.}
\label{tab:results-track-val}
\setlength\tabcolsep{1.5pt}
\begin{threeparttable}
    \begin{tabular}{l | c c c c c c }
        \toprule
        \textbf{Method} & \textbf{AMOTA$\uparrow$} & \textbf{AMOTP$\downarrow$} & \textbf{IDS$\downarrow$} & \textbf{Recall$\uparrow$} & \textbf{MOTA$\uparrow$} & \textbf{MOTP$\downarrow$} \\
        \midrule
        QTrack~\cite{yang2022quality} & 0.347 & 1.347 & 944 & 0.426 & 0.309 & 0.722 \\
        DORT~\cite{qing2023dort} & 0.424 & 1.264 & - & 0.492 & - & - \\
        Sparse4Dv3\textsuperscript{*}~\cite{lin2023sparse4dv3} & 0.490 & 1.164 & 430 & 0.574 & 0.436 & 0.660 \\
        Sparse4Dv3\textsuperscript{\textdaggerdbl}~\cite{lin2023sparse4dv3} & 0.440 & 1.184 & 403 & 0.525 & 0.401 & 0.622 \\
        \rowcolor{Gray} \net & \textbf{0.513} & \textbf{1.109} & \textbf{473} & \textbf{0.595} & \textbf{0.457} & \textbf{0.583} \\
        \bottomrule
    \end{tabular}
    \footnotesize
    All methods use ResNet-50 as the image encoder with an input image resolution of  $256 \times 704$.
    \textdaggerdbl:~Baselines trained with the code provided by the authors.
    *:~Results reported in the original paper but not reproducible.
\end{threeparttable}
\vspace{-0.3cm}
\end{table}

\begin{figure*}
    \centering

    \subfloat[Sparse4Dv3 (VoVNet-99)]{
        \includegraphics[width=0.8\textwidth]{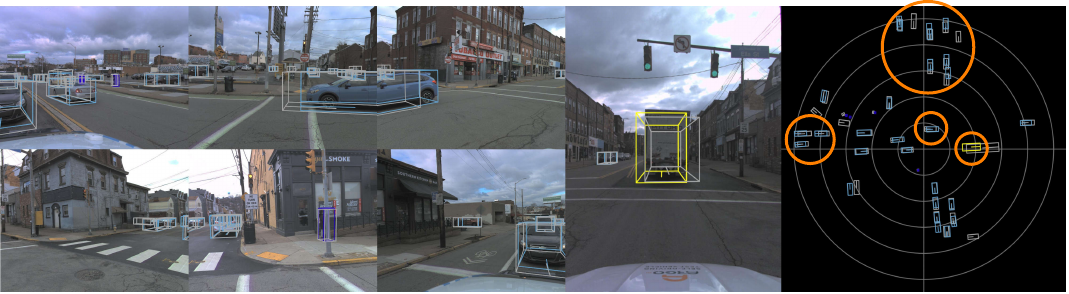}
    }
    \vspace{-.2cm}

    \subfloat[\net (VoVNet-99)]{
        \includegraphics[width=0.8\textwidth]{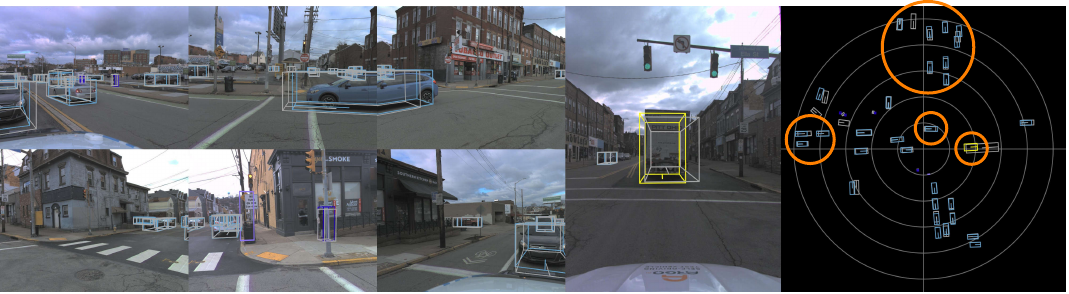}
    }

    \caption{
    Visualization of 3D object detection results
    on the Argoverse~2 val set. Our method shows substantially more accurate object box localization (orange) compared to Sparse4Dv3. Ground truths are colored in gray while predictions are colored by class:\;
    \textcolor[rgb]{0.384,0.635,0.808}{\rule{0.25cm}{0.25cm}} regular vehicle,\;
    \textcolor[rgb]{0.125,0.000,0.757}{\rule{0.25cm}{0.25cm}} pedestrian,\;
    \textcolor[rgb]{1.000,1.000,0.000}{\rule{0.25cm}{0.25cm}} box truck.
    }
    \vspace{-0.3cm}
    
    \label{fig:qualitative-av2}
\end{figure*}

We benchmark \net against strong camera-only baselines across datasets and ranges to assess the impact of augmenting camera-only inference with a retrieved static point cloud prior. On Argoverse~2, we compare at $\SI{150}{\meter}$ and $\SI{50}{\meter}$ ranges (\cref{tab:results-argoverse2-150m} and \cref{tab:results-argoverse2-50m}), and on nuScenes we report detection results in Tab.~\ref{tab:results-3d-val} and tracking results in \cref{tab:results-track-val}. For fairness, we match backbones and image resolutions to the respective baselines. Across all settings, \net consistently improves the primary detection metrics (CDS/NDS and mAP) over our main baseline Sparse4Dv3, and we observe high gains in box localization accuracy (lower mATE). These results align with our motivation: prior static geometry provides metric anchors that reduce depth ambiguity and directly translate into improved 3D object localization. On nuScenes tracking, \net also improves AMOTA and reduces localization error along tracks (lower AMOTP), indicating that geometric priors can benefit not only detection but also object tracking.


\subsection{Ablation Studies}

\begin{table*}[]
\footnotesize
\centering
\caption{Ablation study: Map encoding, fusion, and image-map grid masking.}
\label{tab:ablation-bev-map}
\setlength\tabcolsep{2.5pt}
\begin{threeparttable}
    \begin{tabular}{c c c c | c c | c c | c | c c c}
        \toprule
        \multicolumn{4}{c|}{\textbf{PV Map Encodings}} & \multicolumn{2}{c|}{\textbf{BEV Map Encoding}} & \multicolumn{2}{c|}{\textbf{Camera-Map Fusion}} & \\
        \textbf{Depth} & \textbf{Depth Embed.} & \textbf{Spread} & \textbf{XYZ} & \textbf{Pillar} & \textbf{Voxel} & \textbf{PV Fusion} & \textbf{BEV Fusion} & \textbf{Grid Mask} & \textbf{CDS$\uparrow$} & \textbf{mAP$\uparrow$} & \textbf{mATE$\downarrow$} \\
        \midrule
        & & & & & & & & & 0.289 & 0.378 & 0.698 \\
        \checkmark & & & & & & \checkmark & & & 0.282 & 0.374 & 0.681 \\
        \checkmark & \checkmark & & & & & \checkmark & & & 0.284 & 0.373 & 0.680 \\
        \checkmark & & \checkmark & & & & \checkmark & & & 0.293 & 0.382 & 0.675 \\
        \checkmark & \checkmark & \checkmark & \checkmark & & & \checkmark & & & 0.294 & 0.385 & 0.666 \\
        \midrule
        & & & & \checkmark & & & & & 0.297 & 0.386 & 0.660 \\
        & & & & & \checkmark & & & & 0.301 & 0.394 & 0.645 \\
        \midrule
        \checkmark & \checkmark & \checkmark & \checkmark & & \checkmark & \checkmark & & & 0.287 & 0.378 & 0.658\\
        \checkmark & \checkmark & \checkmark & \checkmark & & \checkmark & \checkmark & \checkmark & & 0.309 & 0.401 & 0.648\\
        \rowcolor{Gray} \checkmark & \checkmark & \checkmark & \checkmark & & \checkmark & \checkmark & \checkmark & \checkmark & 0.309 & 0.401 & 0.645 \\
        \bottomrule
    \end{tabular}
    \footnotesize
     The baseline is Sparse4Dv3 with an image resolution of $480 \times 704$, VoVNet-99 pretrained with FCOS3D as the backbone, and with 3D/BEV data augmentations.
\end{threeparttable}
\vspace{-0.2cm}
\end{table*}

\begin{table}
\footnotesize
\centering
\caption{Results on subsets with higher prior-map coverage.}
\label{tab:results-argoverse2-map-coverage-subsets-net-baseline}
\setlength\tabcolsep{2.0pt}
\begin{threeparttable}
    \begin{tabular}{l | c c c | c c c}
        \toprule
        \multirow{2}{*}{\textbf{Subset}} &
        \multicolumn{3}{c|}{\textbf{\net (map prior)}} &
        \multicolumn{3}{c}{\textbf{Sparse4Dv3 (no map)}} \\
        \cmidrule(lr){2-4}\cmidrule(lr){5-7}
        & \textbf{CDS$\uparrow$} & \textbf{mAP$\uparrow$} & \textbf{mATE$\downarrow$}
        & \textbf{CDS$\uparrow$} & \textbf{mAP$\uparrow$} & \textbf{mATE$\downarrow$} \\
        \midrule
        All & 0.309 {\color{ForestGreen}(+$2.0$\,pp)} & 0.401 & 0.645 & 0.289 & 0.378 & 0.698 \\
        Overlap $k{=}1$ & 0.345 {\color{ForestGreen}(+$3.0$\,pp)} & 0.443 & 0.586 & 0.315 & 0.408 & 0.636 \\
        Overlap $k{=}2$ & 0.336 {\color{ForestGreen}(+$3.7$\,pp)} & 0.429 & 0.607 & 0.299 & 0.388 & 0.670 \\
        \bottomrule
    \end{tabular}
    \footnotesize
    Overlap $k$: a sequence is included iff every $50\,\mathrm{m}\times50\,\mathrm{m}$ tile it covers is also covered by at least $k$ \emph{other} sequences. The map-free baseline is evaluated on the same subsets to isolate the effect of the prior. Gains (green) are in percentage points (pp) for CDS.
\end{threeparttable}
\end{table}

\begin{table}
\footnotesize
\centering
\caption{Robustness to complete map drop at test time.}
\label{tab:results-argoverse2-map-drop-robustness}
\setlength\tabcolsep{0.9pt}
\begin{threeparttable}
    \begin{tabular}{l | c c c c c}
        \toprule
        \textbf{Method} &
        \textbf{CDS$\uparrow$} & \textbf{mAP$\uparrow$} &
        \textbf{mATE$\downarrow$} & \textbf{mASE$\downarrow$} & \textbf{mAOE$\downarrow$} \\
        \midrule
        \net~-- Grid Mask & 0.278 & 0.368 & 0.710 & 0.293 & 0.622 \\
        \rowcolor{Gray} \net &  0.286 & 0.375 & 0.721 & 0.293 & 0.534 \\
        Sparse4Dv3 (map-free) & 0.289 & 0.378 & 0.698 & 0.290 & 0.556 \\
        \bottomrule
    \end{tabular}

    \footnotesize
    For \net variants, the prior-map input is removed at test time. The map-free baseline is unaffected. Grid Mask denotes applying grid-masking augmentation to the map input during training to discourage overreliance on the prior map.
\end{threeparttable}
\vspace{-0.4cm}
\end{table}

We perform controlled ablations on Argoverse~2 to isolate which components are essential for effectively exploiting map priors in~\cref{tab:ablation-bev-map} and to assess robustness under imperfect map availability. Unless stated otherwise, all ablations use VoVNet-99~\cite{lee2019energy} pretraining with FCOS3D with image size $480\times704$ and are evaluated on the Argoverse~2 validation split within a range of 50 meters. We highlight our final configuration in~\cref{tab:ablation-bev-map} and~\cref{tab:results-argoverse2-map-drop-robustness} in gray.\looseness=-1

{\parskip=2pt
\noindent\textit{PV map encodings:}
We first study PV-only fusion by projecting the map onto each camera and concatenating the PV map features with image features, as proposed in~\cite{you2024better}. A naïve PV depth input (depth + mask) does not yield a benefit on our used dataset, revealing that simply concatenating sparse depth is insufficient. Adding depth positional embeddings provides a small improvement, while introducing nearest-valid depth spreading substantially improves CDS/mAP and reduces mATE, indicating that making sparse geometry spatially usable in PV is critical. Augmenting the PV encoding with ego-frame coordinate channels $(x,y,z)$ further improves box localization.\looseness=-1

{\parskip=2pt
\noindent\textit{BEV map encoding:}
We evaluate BEV-only map usage by encoding the retrieved point cloud in BEV and attending to it in the head. Both pillar-based~\cite{yuan2024presight} and sparse-voxel~\cite{yan2018second} encoders are effective, but sparse voxel encoding achieves stronger results.

{\parskip=2pt
\noindent\textit{Where to fuse: PV vs.\ BEV vs.\ both:}
Combining PV and BEV branches is not a trivial union of the best single-branch variants. Simply enabling both can underperform due to representation mismatch and training dynamics. In contrast, fusing in PV \emph{and} explicitly lifting the PV-fused features to BEV for BEV fusion yields the strongest improvements, showing that BEV fusion in a shared metric space is essential to fully exploit map priors. This dual-space design also distinguishes our approach from PV-only prior-map fusion~\cite{you2024better} by enabling metric neighborhood reasoning in BEV while retaining PV object-centric details via PV deformable aggregation.

{\parskip=2pt
\noindent\textit{Map coverage and robustness:}
Not all validation sequences have a map prior. To quantify the dependence on coverage, we evaluate subsets of sequences with higher map overlap in~\cref{tab:results-argoverse2-map-coverage-subsets-net-baseline}, while also evaluating the map-free baseline on the same subsets to isolate the effect of the prior. Performance gains grow with coverage, confirming that \net is most effective when reliable priors are available. Finally, we evaluate robustness to complete map drop at test time in ~\cref{tab:results-argoverse2-map-drop-robustness}. Without map-specific regularization, performance can fall below the map-free baseline, indicating overreliance on priors. Applying image-map grid masking during training makes our \net competitive with the map-free baseline even when the prior is entirely missing at test time, while preserving the benefits when the map is available.


\subsection{Qualitative Results}

\cref{fig:qualitative-av2} presents qualitative comparisons of our \net with Sparse4Dv3, in both PV and BEV on Argoverse~2. In challenging scenarios with clutter and many occlusions, we observe that \net yields tighter 3D box localization and more precise box predictions along the depth for both distant and close objects. We also observe that objects such as cars align more closely with the lane layout. This supports our quantitative finding that leveraging prior static geometry reduces depth ambiguity and improves 3D object localization.

\section{Conclusion}
In this paper, we presented DualViewMapDet, a novel method that improves camera-based 3D object detection and tracking on nuScenes and Argoverse~2 datasets by leveraging a clever data flow and eliminating the need for precise LiDAR depth information at inference time. For data flow, we proposed enriching both PV and BEV features with a PV-projected and BEV-encoded prior map. By fusing the map-enriched PV features lifted to BEV with map-enriched BEV features, we achieved state-of-the-art performance on both tasks without access to LiDAR-precise depth during the drive. We believe that overcoming the limitations of one-sided data flows and encoding precise 3D geometry into such systems will increase their performance and reliability.

\balance
\footnotesize
\bibliographystyle{IEEEtran}
\bibliography{references.bib}

@String { advneurips           = {Advances in Neural Information Processing Systems} }

@String { CVPR                 = {{IEEE/CVF} Conference on Computer Vision and Pattern Recognition} }

@String { ECCV                 = {European Conference on Computer Vision} }

@String { ICCV                 = {International Conference on Computer Vision} }

@String { ICLR                 = {International Conference on Learning Representations}}

@String { ieeeral              = {{IEEE} Robotics and Automation Letters} }

@String { CORL                 = {Conference on Robot Learning} }

@String { ieeeral              = {Robotics and Automation Letters} }

@String { AAAI                   = {AAAI Conference on Artificial Intelligence} }

@String { IROS                 = {{IEEE/RSJ} International Conference on Intelligent Robots and Systems}}

@String { ICRA                 = {{IEEE} International Conference on Robotics and Automation}}

@article{yan2018second,
  title={Second: Sparsely embedded convolutional detection},
  author={Yan, Yan and Mao, Yuxing and Li, Bo},
  journal={Sensors},
  volume={18},
  number={10},
  pages={3337},
  year={2018},
}

@inproceedings{yin2021center,
  title={Center-based 3d object detection and tracking},
  author={Yin, Tianwei and Zhou, Xingyi and Krahenbuhl, Philipp},
  booktitle=CVPR,
  pages={11784--11793},
  year={2021}
}

@inproceedings{yang2023bevformer,
  author={Yang, Chenyu and Chen, Yuntao and Tian, Hao and Tao, Chenxin and Zhu, Xizhou and Zhang, Zhaoxiang and others},
  booktitle=CVPR,
  title={Bevformer v2: Adapting modern image backbones to bird's-eye-view recognition via perspective supervision},
  year={2023},
  pages={17830--17839},
}

@inproceedings{philion2020lift,
  title={Lift, splat, shoot: Encoding images from arbitrary camera rigs by implicitly unprojecting to 3d},
  author={Philion, Jonah and Fidler, Sanja},
  booktitle=ECCV,
  pages={194--210},
  year={2020},
}

@inproceedings{li2023bevdepth,
  title={Bevdepth: Acquisition of reliable depth for multi-view 3d object detection},
  author={Li, Yinhao and Ge, Zheng and Yu, Guanyi and Yang, Jinrong and Wang, Zengran and Shi, Yukang and Sun, Jianjian and Li, Zeming},
  booktitle=AAAI,
  year={2023}
}

@inproceedings{li2024bevnext,
  title={BEVNeXt: Reviving Dense BEV Frameworks for 3D Object Detection},
  author={Li, Zhenxin and Lan, Shiyi and Alvarez, Jose M and Wu, Zuxuan},
  booktitle=CVPR,
  pages={20113--20123},
  year={2024}
}

@inproceedings{liu2022petr,
  title={Petr: Position embedding transformation for multi-view 3d object detection},
  author={Liu, Yingfei and Wang, Tiancai and Zhang, Xiangyu and Sun, Jian},
  booktitle=ECCV,
  pages={531--548},
  year={2022},
}

@inproceedings{wang2023exploring,
  title={Exploring object-centric temporal modeling for efficient multi-view 3d object detection},
  author={Wang, Shihao and Liu, Yingfei and Wang, Tiancai and Li, Ying and Zhang, Xiangyu},
  booktitle=ICCV,
  pages={3621--3631},
  year={2023}
}

@article{lin2023sparse4dv3,
  title={Sparse4d v3: Advancing end-to-end 3d detection and tracking},
  author={Lin, Xuewu and Pei, Zixiang and Lin, Tianwei and others},
  journal={arXiv preprint arXiv:2311.11722},
  year={2023}
}

@article{kappeler2025bridging,
  title={Bridging Perspectives: Foundation Model Guided BEV Maps for 3D Object Detection and Tracking},
  author={K{\"a}ppeler, Markus and {\c{C}}i{\c{c}}ek, {\"O}zg{\"u}n and Cattaneo, Daniele and Gl{\"a}ser, Claudius and Miron, Yakov and Valada, Abhinav},
  journal={arXiv preprint arXiv:2510.10287},
  year={2025}
}

@inproceedings{jiang2024far3d,
  title={Far3d: Expanding the horizon for surround-view 3d object detection},
  booktitle=AAAI,
  year={2024}
}

@inproceedings{liu2024ray,
  title={Ray denoising: Depth-aware hard negative sampling for multi-view 3d object detection},
  author={Liu, Feng and Huang, Tengteng and Zhang, Qianjing and Yao, Haotian and Zhang, Chi and Wan, Fang and Ye, Qixiang and Zhou, Yanzhao},
  booktitle=ECCV,
  pages={200--217},
  year={2024},
}

@article{han2024exploring,
  title={Exploring recurrent long-term temporal fusion for multi-view 3d perception},
  author={Han, Chunrui and Yang, Jinrong and Sun, Jianjian and Ge, Zheng and Dong, Runpei and Zhou, Hongyu and Mao, Weixin and Peng, Yuang and Zhang, Xiangyu},
  journal=ieeeral,
  year={2024},
  volume={9},
  number={7},
  pages={6544-6551},
}

@inproceedings{bai2022transfusion,
  title={Transfusion: Robust lidar-camera fusion for 3d object detection with transformers},
  author={Bai, Xuyang and Hu, Zeyu and Zhu, Xinge and Huang, Qingqiu and Chen, Yilun and Fu, Hongbo and Tai, Chiew-Lan},
  booktitle={Proceedings of the IEEE/CVF conference on computer vision and pattern recognition},
  pages={1090--1099},
  year={2022}
}

@article{liu2022bevfusion,
  title={Bevfusion: Multi-task multi-sensor fusion with unified bird's-eye view representation},
  author={Liu, Zhijian and Tang, Haotian and Amini, Alexander and Yang, Xinyu and Mao, Huizi and Rus, Daniela and Han, Song},
  journal={arXiv preprint arXiv:2205.13542},
  year={2022}
}

@article{cai2023bevfusion4d,
  title={Bevfusion4d: Learning lidar-camera fusion under bird's-eye-view via cross-modality guidance and temporal aggregation},
  author={Cai, Hongxiang and Zhang, Zeyuan and Zhou, Zhenyu and others},
  journal={arXiv preprint arXiv:2303.17099},
  year={2023}
}

@article{hu2023ea,
  title={Ea-lss: Edge-aware lift-splat-shot framework for 3d bev object detection},
  author={Hu, Haotian and Wang, Fanyi and Su, Jingwen and Wang, Yaonong and Hu, Laifeng and Fang, Weiye and Xu, Jingwei and Zhang, Zhiwang},
  journal={arXiv preprint arXiv:2303.17895},
  year={2023}
}

@inproceedings{yin2024fusion,
  title={Is-fusion: Instance-scene collaborative fusion for multimodal 3d object detection},
  author={Yin, Junbo and Shen, Jianbing and Chen, Runnan and Li, Wei and Yang, Ruigang and Frossard, Pascal and Wang, Wenguan},
  booktitle=CVPR,
  pages={14905--14915},
  year={2024}
}

@inproceedings{song2024graphbev,
  title={Graphbev: Towards robust bev feature alignment for multi-modal 3d object detection},
  author={Song, Ziying and Yang, Lei and Xu, Shaoqing and Liu, Lin and Xu, Dongyang and Jia, Caiyan and Jia, Feiyang and Wang, Li},
  booktitle=ECCV,
  year={2024},
}

@inproceedings{mohan2024progressive,
  title={Progressive multi-modal fusion for robust 3d object detection},
  author={Mohan, Rohit and Cattaneo, Daniele and Drews, Florian and Valada, Abhinav},
  booktitle=CORL,
  year={2024}
}

@inproceedings{qing2023dort,
  title={Dort: Modeling dynamic objects in recurrent for multi-camera 3d object detection and tracking},
  author={Qing, LIAN and Wang, Tai and Lin, Dahua and Pang, Jiangmiao},
  booktitle=CORL,
  pages={3749--3765},
  year={2023},
}

@article{yang2022quality,
  title={Quality matters: Embracing quality clues for robust 3d multi-object tracking},
  author={Yang, Jinrong and Yu, En and Li, Zeming and Li, Xiaoping and Tao, Wenbing},
  journal={arXiv preprint arXiv:2208.10976},
  year={2022}
}

@inproceedings{yuan2024presight,
  title={Presight: Enhancing autonomous vehicle perception with city-scale nerf priors},
  author={Yuan, Tianyuan and Mao, Yucheng and Yang, Jiawei and Liu, Yicheng and Wang, Yue and Zhao, Hang},
  booktitle=ECCV,
  pages={323--339},
  year={2024},
}

@article{yuan2025lmpocc,
  title={LMPOcc: 3D Semantic Occupancy Prediction Utilizing Long-Term Memory Prior from Historical Traversals},
  author={Yuan, Shanshuai and Wei, Julong and Tie, Muer and Ren, Xiangyun and Gan, Zhongxue and Ding, Wenchao},
  journal={arXiv preprint arXiv:2504.13596},
  year={2025}
}

@inproceedings{yuan2025collaborative,
  title={Collaborative Perceiver: Elevating Vision-Based 3D Object Detection via Local Density-Aware Dense Spatial Occupancy},
  author={Yuan, Jicheng and Nguyen-Duc, Manh and Liu, Qian and Hauswirth, Manfred and Le-Phuoc, Danh},
  booktitle={International Conference on Neural Information Processing},
  pages={48--62},
  year={2025},
}

@article{yu2024ultimatedo,
  title={UltimateDO: An Efficient Framework to Marry Occupancy Prediction with 3D Object Detection via Channel2height},
  author={Yu, Zichen and Shu, Changyong},
  journal={arXiv preprint arXiv:2409.11160},
  year={2024}
}

@article{li2024memorize,
  title={Memorize what matters: Emergent scene decomposition from multitraverse},
  author={Li, Yiming and Wang, Zehong and Wang, Yue and Yu, Zhiding and Gojcic, Zan and Pavone, Marco and Feng, Chen and Alvarez, Jose M},
  journal=advneurips,
  volume={37},
  pages={108389--108438},
  year={2024}
}

@inproceedings{you2022hindsight,
  title = {Hindsight is 20/20: Leveraging Past Traversals to Aid 3D Perception},
  author = {You, Yurong and Luo, Katie Z and Chen, Xiangyu and Chen, Junan and Chao, Wei-Lun and Sun, Wen and Hariharan, Bharath and Campbell, Mark and Weinberger, Kilian Q.},
  booktitle = ICLR,
  year = {2022},
}

@inproceedings{you2024better,
  title={Better monocular 3d detectors with lidar from the past},
  author={You, Yurong and Phoo, Cheng Perng and Diaz-Ruiz, Carlos Andres and Luo, Katie Z and Chao, Wei-Lun and Campbell, Mark and Hariharan, Bharath and Weinberger, Kilian Q},
  booktitle=ICRA,
  pages={6634--6641},
  year={2024},
}

@article{sun2024sparsedrive,
  title={SparseDrive: End-to-End Autonomous Driving via Sparse Scene Representation},
  author={Sun, Wenchao and Lin, Xuewu and Shi, Yining and Zhang, Chuang and Wu, Haoran and Zheng, Sifa},
  journal={arXiv preprint arXiv:2405.19620},
  year={2024}
}

@article{caesar2020nuscenes,
  title={Panoptic nuscenes: A large-scale benchmark for lidar panoptic segmentation and tracking},
  author={Fong, Whye Kit and Mohan, Rohit and Hurtado, Juana Valeria and Zhou, Lubing and Caesar, Holger and Beijbom, Oscar and Valada, Abhinav},
  journal=ieeeral,
  volume={7},
  number={2},
  pages={3795--3802},
  year={2022},
  publisher={IEEE}
}

@INPROCEEDINGS {wilson2023argoverse,
  author = {Benjamin Wilson and William Qi and Tanmay Agarwal and John Lambert and Jagjeet Singh and Siddhesh Khandelwal and others},
  title = {Argoverse 2: Next Generation Datasets for Self-driving Perception and Forecasting},
  booktitle = {Proc. of the Neural Information Processing Systems Track on Datasets and Benchmarks},
  year = {2021}
}

@article{zhou2018open3d,
  title={Open3D: A modern library for 3D data processing},
  author={Zhou, Qian-Yi and Park, Jaesik and Koltun, Vladlen},
  journal={arXiv preprint arXiv:1801.09847},
  year={2018}
}

@article{vaswani2017attention,
  title={Attention is all you need},
  author={Vaswani, Ashish and Shazeer, Noam and Parmar, Niki and Uszkoreit, Jakob and Jones, Llion and Gomez, Aidan N and Kaiser, {\L}ukasz and Polosukhin, Illia},
  journal=advneurips,
  volume={30},
  year={2017}
}

@article{cattaneo2025cmrnext,
  title={CMRNext: Camera to LiDAR matching in the wild for localization and extrinsic calibration},
  author={Cattaneo, Daniele and Valada, Abhinav},
  journal={IEEE Transactions on Robotics},
  year={2025},
}

@inproceedings{he2016deep,
  title={Deep residual learning for image recognition},
  author={He, Kaiming and Zhang, Xiangyu and Ren, Shaoqing and Sun, Jian},
  booktitle=CVPR,
  pages={770--778},
  year={2016}
}

@inproceedings{lee2019energy,
  title={An energy and GPU-computation efficient backbone network for real-time object detection},
  author={Lee, Youngwan and Hwang, Joong-won and Lee, Sangrok and Bae, Yuseok and Park, Jongyoul},
  booktitle={Proc. of the IEEE/CVF conf. on computer vision and pattern recognition workshops},
  year={2019}
}

@article{chen2020gridmask,
  title={Gridmask data augmentation},
  author={Chen, Pengguang and Liu, Shu and Zhao, Hengshuang and Wang, Xingquan and Jia, Jiaya},
  journal={arXiv preprint arXiv:2001.04086},
  year={2020}
}

@inproceedings{lim2025kiss,
  title={Kiss-matcher: Fast and robust point cloud registration revisited},
  author={Lim, Hyungtae and Kim, Daebeom and Shin, Gunhee and Shi, Jingnan and Vizzo, Ignacio and Myung, Hyun and Park, Jaesik and Carlone, Luca},
  booktitle=ICRA,
  pages={11104--11111},
  year={2025},
}

@article{buchner20223d,
  title={3d multi-object tracking using graph neural networks with cross-edge modality attention},
  author={B{\"u}chner, Martin and Valada, Abhinav},
  journal=ieeeral,
  volume={7},
  number={4},
  pages={9707--9714},
  year={2022},
}

@inproceedings{nallapareddy2023evcenternet,
  title={EvCenterNet: Uncertainty estimation for object detection using evidential learning},
  author={Nallapareddy, Monish R and Sirohi, Kshitij and Drews, Paulo LJ and Burgard, Wolfram and Cheng, Chih-Hong and Valada, Abhinav},
  booktitle=IROS,
  pages={5699--5706},
  year={2023},
}

@inproceedings{lang2024point,
  title={A point-based approach to efficient lidar multi-task perception},
  author={Lang, Christopher and Braun, Alexander and Schillingmann, Lars and Valada, Abhinav},
  booktitle=IROS,
  year={2024},
}

@article{gosala2026sparse3dtrack,
  title={Sparse3DTrack: Monocular 3D Object Tracking Using Sparse Supervision},
  author={Gosala, Nikhil and Kiran, B Ravi and Yogamani, Senthil and Valada, Abhinav},
  journal={arXiv preprint arXiv:2603.18298},
  year={2026}
}


\end{document}